\documentclass{bmvc2k}

\title{Temporal Feature Warping for Video Shadow Detection}

\addauthor{Shilin Hu}{shilhu@cs.stonybrook.edu}{1}
\addauthor{Hieu Le}{hle@cs.stonybrook.edu}{1}
\addauthor{Dimitris Samaras}{samaras@cs.stonybrook.edu}{1}

\addinstitution{
 Computer Science Department\\
 Stony Brook University\\
 New York, USA
}

\runninghead{Hu \etal}{Temporal Feature warping for VSD}

\def\eg{\emph{e.g}\bmvaOneDot}

\def\etal{\emph{et al}\bmvaOneDot}

\begin{document}
\maketitle
\begin{abstract}
While single image shadow detection has been improving rapidly in recent years, video shadow detection remains a challenging task due to  data scarcity and the difficulty in modelling temporal consistency. 
The current video shadow detection method achieves this goal via co-attention, which mostly exploits information that is temporally coherent but is not robust in detecting moving shadows and small shadow regions.
In this paper, we propose a simple but powerful method to better aggregate information temporally.
We use an optical flow based warping module to align and then combine features between frames.
We apply this warping module across multiple deep-network layers to retrieve information from neighboring frames including both local details and high-level semantic information. 
We train and test our framework on the ViSha dataset.
Experimental results show that our model outperforms the state-of-the-art video shadow detection  method by 28\%, reducing BER from 16.7 to 12.0.
\end{abstract}

\section{Introduction}
\label{sec:intro}
Shadows appear in most natural images. Simply detecting shadows benefits many computer vision tasks such as image classification \cite{Filippi_13}, image segmentation \cite{XU2019142} and object tracking \cite{Saravanakumar_2010, Mohanapriya_ICICES_2017}.
Therefore, shadow detection has drawn a lot of interest in recent years especially with the rapid development of deep-learning-based methods \cite{Nguyen_2017_ICCV, Zheng_2019_CVPR, Wang_2018_CVPR, Ding_2019_ICCV, Le_2018_ECCV}.
However, recent shadow detection works mostly deal with shadows in single images while video shadow detection remains an open question despite many potential applications \cite{Wang2020PeopleAS,Le_2020_ECCV, le2020physicsbased, Le2016GeodesicDH}. 

\begin{figure}
\includegraphics[width=\textwidth]{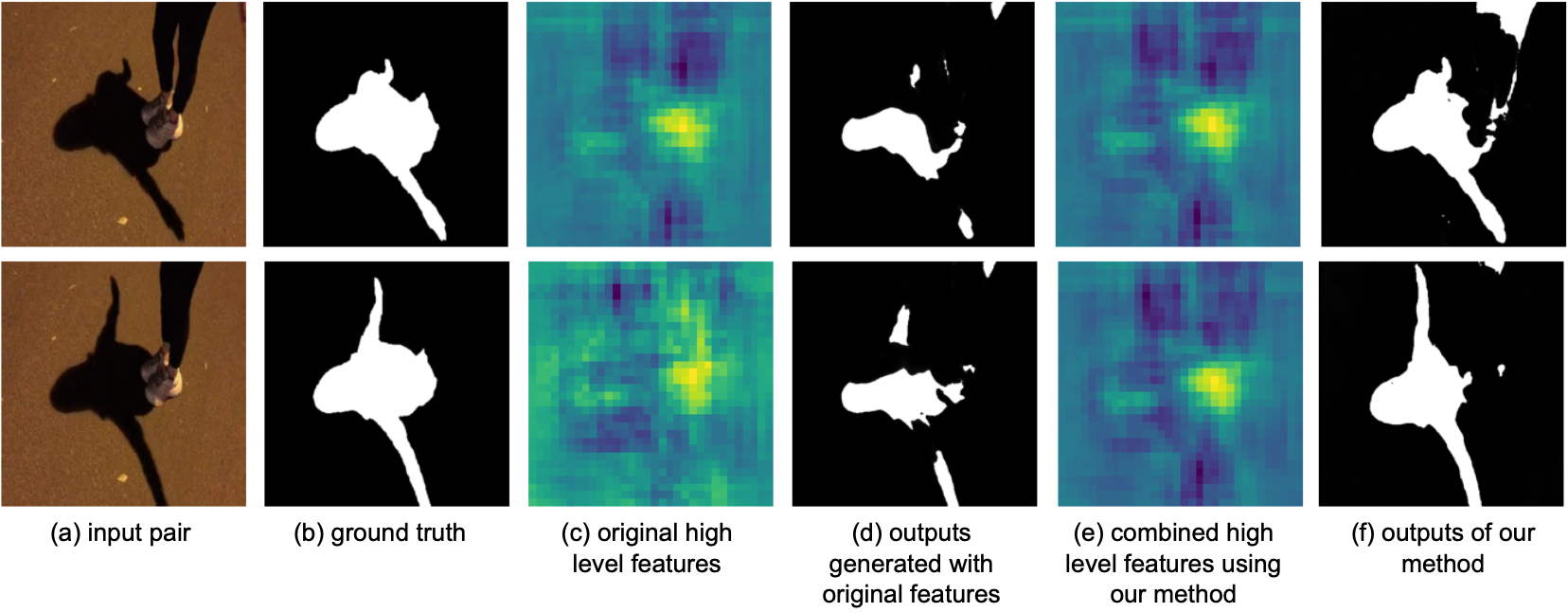}
\caption{\textbf{Motivation of our work.} (a) shows a pair of consecutive video frames, (b) shows the ground truth shadow masks, (c) shows the original extracted high level feature maps. Even though the difference between images is small, the difference between features is high, (d) shows shadow mask prediction using original features, (e) is combined features obtained by our method, (f) shows shadow prediction using combined features. Here we demonstrate the effectiveness of our method; the combined feature maps preserve correspondence between frames and produce more stable predictions.}
\label{fig:1}
\end{figure}

A shadow video typically consists of hundreds of frames that contain shadows varying in shapes and intensities. The detection problem is compounded by video-specific issues such as motion blur. Thus, simply applying image shadow detection methods frame-by-frame on video often yields inconsistent predictions (see Fig \ref{fig:1}.d). 
Instead, a common strategy to deal with video data is to leverage temporal information across video frames \cite{Zhu_2017_CVPR, Yan_2019_ICCV}. Here the difficulty lies in how to incorporate information across video where there is spatial misalignment between frames due to the movements of objects and cameras. 

In this paper, we propose a straightforward but powerful deep-learning based method to obtain a rich feature representation for video shadow detection. We focus on dealing with the temporal misalignment between frame representations. Our strategy is simple: we align the features across frames by optical flow and then linearly combine them to obtain the per-frame final feature representation. Optical flow is easy to obtain \cite{Liu_2020_CVPR} and can effectively align spatial content, including small details. As this warping is computationally efficient, we can apply it on multiple layers of our network.

We report that this simple optical-flow based feature aggregation scheme works surprisingly well for shadow detection in videos. We train our model in an end-to-end fashion on the Visha dataset \cite{chen21TVSD}. Our method achieves state-of-the-art video shadow detection performance, outperforming the previous method \cite{chen21TVSD} by a 28\% BER reduction. Fig \ref{fig:1} illustrates the effect of our proposed temporal feature warping method. As can be seen, the features of two visually similar frames could be wildly different (column c), which results in inconsistent outputs (column d). Our method warps and then combines the features, making them consistent across frames (column e), and finally outputs stable and temporally consistent results (last column). 

\section{Related Work}
\label{sec:related}
Single image shadow detection is a well studied topic. On one hand, earlier research on image shadow detection mostly focuses on spectral or spatial features of images such as chromaticity, physical properties, geometry, and texture \cite{SANIN20121684}. 
On the other hand, recent shadow detection methods show tremendous success with the rapid development of deep learning. Le \etal \cite{Le_2018_ECCV} propose to train a shadow detection network together with a shadow attenuation network that generates adversarial training examples. 
Hu \etal \cite{Hu_2018_CVPR} propose a directional-aware feature extractor for aggregating spatial information. Zhu \etal \cite{Zhu_2018_ECCV} utilize recurrent attention residual modules to fully aggregate the global and local contexts in different layers of the CNN to detect shadows. Chen \etal \cite{Chen_2020_CVPR} further improve detection performance by introducing a multi-task mean teacher architecture which leverages unlabeled data. 
However, image methods trained on image datasets such as SBU \cite{vicente2016large} and CUHK-Shadow \cite{hu2021revisiting} do not generalize well to videos due to the lack of temporal consistency. 

Video shadow detection is a classic problem on its own. Earlier work \cite{Jacques_SIBGRAPI_2005, Shi_ICMLA_2019} focuses on the spectral and spatial features, which depend heavily  on the quality of data. Without temporal constraints, these methods often output inconsistent predictions across frames.
The first large-scale video shadow detection dataset was proposed by
Chen \etal \cite{chen21TVSD}. The dataset contains  120 fully-annotated videos with a total of 11,685 frames. They also proposed the first deep-learning-based method for video shadow detection in which a dual gated co-attention module is used to focus on common high-level features between frames. This co-attention module allows their network to filter out temporally inconsistent information from each frame representation to obtain more stable and consistent results.
However, this mechanism causes the method to be less sensitive to shadow areas that substantially change across frames due to temporal misalignment. By contrast, our method performs temporal alignment before combining features. Besides, our temporal alignment module can be used for all layers of the network, allowing us to pick up even small shadow areas. Co-attention can only be applied on high-level feature maps due to its computational cost. 


Optical flow is used in various high-level video tasks \cite{SHIN2005204, Zhong_Liu_Ren_Zhang_Ren_2013, Buades_2016}. Recent deep learning based methods \cite{Ilg_2017_CVPR, Sun_2018_CVPR} for optical flow estimation are fairly accurate and efficient in inference. However, most popular datasets used in training optical flow estimation models, \eg, MPI Sintel \cite{Butler:ECCV:2012} and KITTI 2015 \cite{Menze2015CVPR}, are sufficiently different from shadow detection datasets \cite{Wang_2018_CVPR,Yago16,hu2021revisiting}. To compensate for this domain shift, we use a simple module to refine optical flow before using it to warp our features.  Liu \etal \cite{Liu_2020_CVPR} propose an unsupervised dense optical flow estimation network with better cross dataset generalization capability by learning from abundant augmentations of training data. 

\section{Method}
\label{sec:method}

\subsection{Overview}

\begin{figure}
\includegraphics[width=\textwidth]{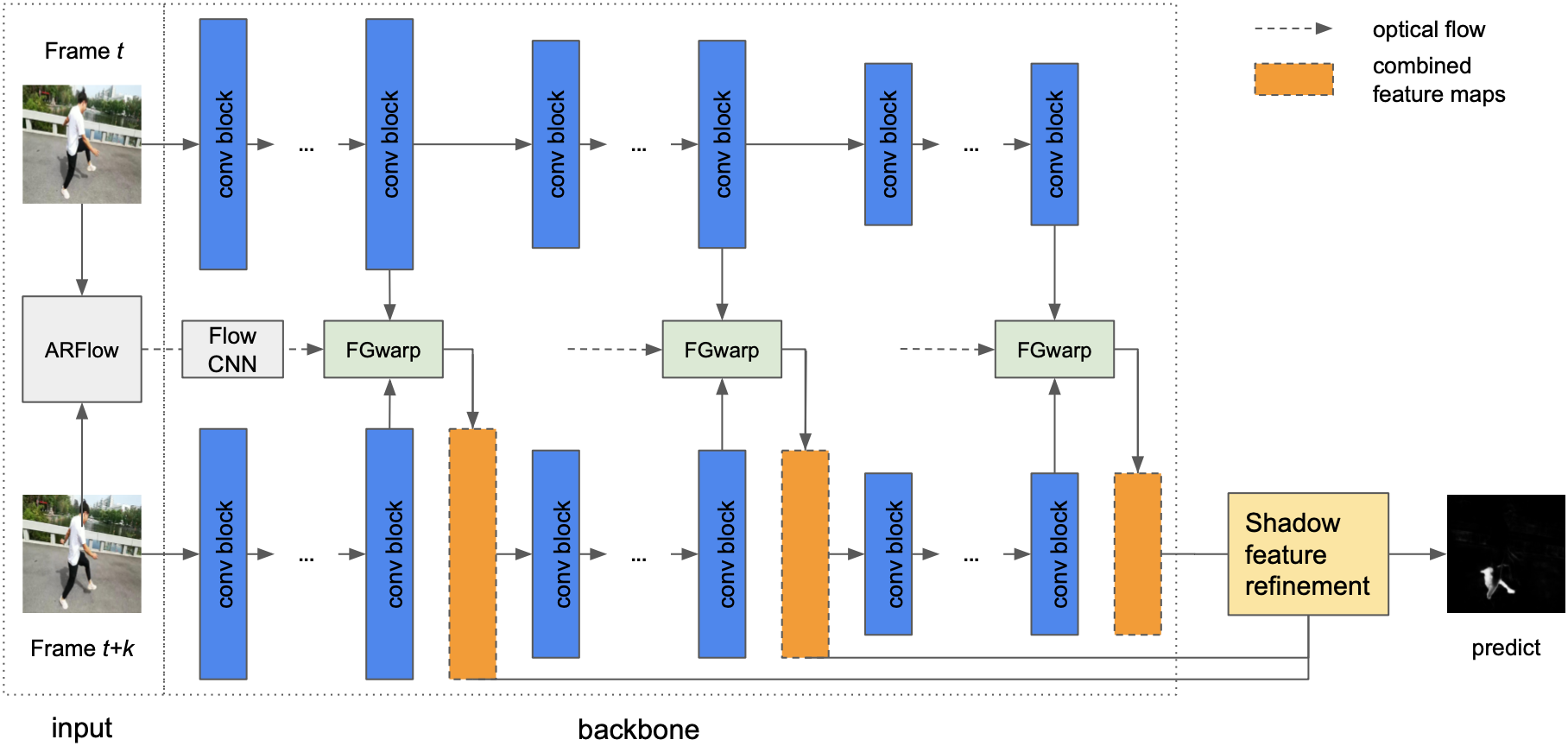}
\caption{\textbf{Simplified network architecture.} Our network consists of two branches with identical architecture. Each branch predicts the shadow mask for the corresponding input frame. Here we only show the data flow from the ``\textit{Frame t}'' branch to the ``\textit{Frame t+k}'' branch for simplicity. In practice, feature warping and combination goes both ways. At multiple layers of the network, we feed the outputs of both branches into a flow-guided warp (FGwarp) module (details in Fig.\ref{fig:3}) to obtain a temporally enriched feature (colored as orange). We use a shadow feature refinement module to predict the shadow mask from all combined features.}
\label{fig:2}
\end{figure}

The overall structure of our framework is illustrated in Fig \ref{fig:2}. Our model consists of two branches with identical architectures. The input of our model is a  RGB video frame pair.
The two images are input to the two branches of the model to extract two sets of feature maps across the three different layers of the network. 
Throughout the network, these features are progressively enriched by the information from the features of the other image via temporal warping and linear combination. 

In particular, we first obtain two dense optical flow fields from the two images using ARFlow \cite{Liu_2020_CVPR}. 
Following \cite{Gadde_2017_ICCV, Li2021}, we train a small module to refine these optical flow fields to better suit the feature warping task, depicted as \textit{FlowCNN} in Fig \ref{fig:2}. At multiple layers of the network, we combine the features of each branch with the aligned features from the other. 
A flow-guided warp (FGwarp) module is used to first warp the frame feature representation to spatially align it with the content of the other frame and then linearly combine the original features with the warped features from the other frame. 
This simple feature aggregation scheme ensures consistency between the two frame representations. The combined features of different levels are input through a shadow refinement module to generate the final shadow mask prediction.

\subsection{Shadow detector network}
Our network consists of two branches with identical architecture.
For each branch, we use the MobileNet V2 \cite{Sandler_2018_CVPR} as the backbone feature extractor. Each branch consists of a series of inverted residual bottlenecks (IRBs) \cite{Sandler_2018_CVPR}.
We input the feature maps of the $4^{th}$, the $7^{th}$ and the last block of each branch, which encode the low, mid, and high-level features respectively, to the \textit{FGwarp} module to obtain the corresponding combined features. 
Finally, these three combined feature maps are input to a detail enhancement module \cite{hu2021revisiting} to refine the features and predict the final shadow mask.

\subsection{Optical flow estimation}
We use a pre-trained ARFlow \cite{Liu_2020_CVPR} model to produce optical flow for our two input frames. This network is trained on MPI Sintel \cite{Butler:ECCV:2012} which differs in object types and occlusions from our shadow video data. Thus, we train a flow refinement module, FlowCNN, to refine (by adapting the domain) the output of ARFlow to better fit the feature warping task. The input for the FlowCNN consists of the optical flow from ARFlow, the two input frames, and the pixel-wise difference of the two frames. The network consists of 4 convolutional layers, in which the first two layers are followed by a BatchNorm and a ReLU layer. The output of the third layer is then concatenated with the original flow and passed to the last convolution layer to obtain the refined optical flow. We train our proposed FlowCNN together with the shadow detector network in an end-to-end manner.

\subsection{Flow warping and combination}

\begin{figure}
\begin{center}
    \includegraphics[width=0.6\textwidth]{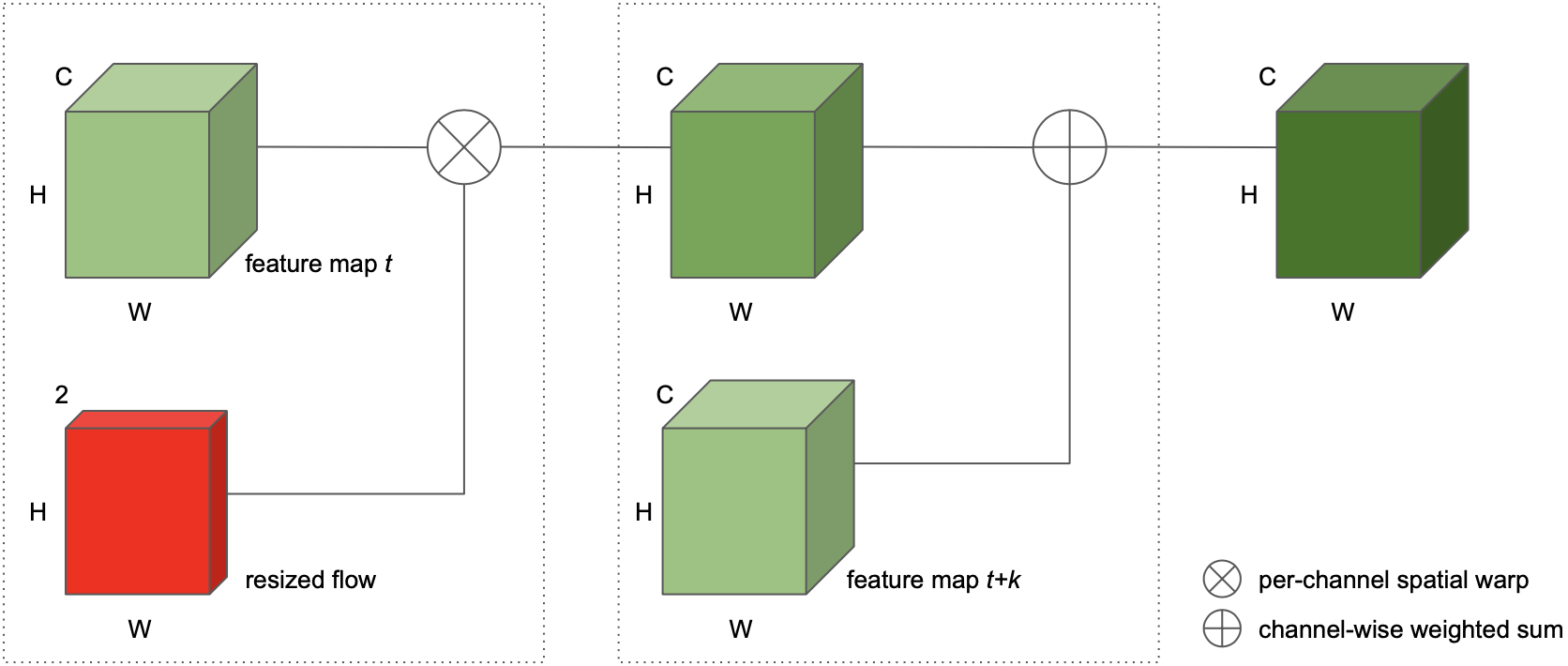}
\end{center}
\caption{\textbf{FGwarp module.} The feature map from frame \textit{t} is warped using the flow field computed from our FlowCNN. The warped feature is then linearly combined with the original feature of frame \textit{t+k} to form the aggregated features. Note that we resize the flow field to the size of the input feature maps.}
\label{fig:3}
\end{figure}

We enforce consistency between frame feature representations by mutually exchanging their intermediate features. Since motion in video causes the spatial misalignment between the content of consecutive frames, we first need to apply an optical flow based feature warping to align the features. A flow-guided warp (FGwarp) module is defined for the task.
Fig \ref{fig:3} illustrates this scheme for transferring features from frame $t$ to frame $t+k$. 

Given a pair of feature maps, $f_1$ and $f_2$, and the refined optical flow $U_{1\rightarrow{2}}$. The feature $f_2$ can be warped to spatially align with the contents of $f_1$. We perform this feature warping for each channel of the feature map separately. The value at a pixel $p$, channel $c$ of the warped feature $\hat{f}_1^c$ can be computed as follows:
\begin{equation}
    \hat{f}_1^c(p) = \sum_q{B(q, p + U_{1\rightarrow{2}}(p))f_2^c(q)}
\label{eq:1}
\end{equation}
where $q$ enumerates all spatial locations in the feature map and $B(\cdot,\cdot)$ denotes the bi-linear interpolation kernel. 

Given the refined optical flow ${U}_{{t}\rightarrow{t+k}}$ and the feature map $f_{t}$, we can propagate the features from the $t^{th}$ frame to get the aligned $\hat{f}_{t+k}$:
\begin{equation}
    \hat{f}_{t+k} = warp(f_{t}, U_{{t}\rightarrow{t+k}}),\\
\label{eq:2}
\end{equation}

The $warp(\cdot, \cdot)$ function is implemented by applying Eq.\ref{eq:1} on different  feature map levels. Then, the combined feature map $\Tilde{f}_{t+k}$ can be computed as follows: 
\begin{equation}
    \Tilde{f}_{t+k} = w_1 \odot f_{t+k} + w_2 \odot \hat{f}_{t+k},\\
\label{eq:3}
\end{equation}
where $w_1$ and $w_2$ represent the per-channel coefficients of the linear combination that combines the two features, $\odot$ represents channel-wise scalar multiplication. The resulting $\Tilde{f}_{t+k}$ is then passed to the following layers in the feature extractor backbone.

\subsection{Training and Inference}
We implement our framework using PyTorch. The feature extractor backbone is initialized via pre-trained MobileNet V2 on ImageNet \cite{Deng_2009_CVPR} while other components are trained from scratch. We use stochastic gradient descent (SGD) with momentum of $0.9$ and weight decay of $0.0005$ to optimize the whole network. The training is in an end-to-end fashion where the objective function is to minimize the mean squared error (MSE) between the ground-truth and predicted shadow masks. The initial learning rate is set to $0.005$, updated by poly strategy \cite{liu2015parsenet} with the power of $0.9$. We train the model for $20$k iterations. All input frames are resized to $512\times 512$. The coefficient vectors $w_1$ and $w_2$ are initialized as $w_1=[1,1..,1]$ and $w_2=[0,0..,0]$, i.e., the temporal warping is not enforced at start. $k$ is set to 1 in training, i.e., adjacent frames are used as input pairs to train our model.

For inference, we resize the inputs to $512\times 512$. We predict shadow masks for each pair of adjacent frames. Thus, each frame, except the first and last one, will be used in two different inference passes. The final shadow mask of each frame is the average of the frame's two output shadow masks. 

\section{Experiments and Results}
\label{sec:exre}

\subsection{Evaluation datasets and metrics}
\textbf{Benchmark dataset} We use the ViSha dataset \cite{chen21TVSD} to evaluate our proposed method. The ViSha dataset has 4788 frames from 50 videos for training and 6897 frames from 70 videos for testing. All methods are trained on the training set and evaluated on the testing set for a fair comparison.

\textbf{Evaluation metric} We employ the commonly-used balanced error rate (BER) to evaluate shadow detection performance, which is defined as: BER $ = 100\times(1-\frac{1}{2}(\frac{TP}{TP+FN}+\frac{TN}{TN+FP}))$, where $TP,TN,FP,FN$ are the total numbers of true positive, true negative, false positive, and false negative pixels respectively.
Since shadow pixels are usually minority in natural images, the BER is less biased than mean pixel accuracy.
In general, lower BER indicates better shadow detection performance. 
We also provide separate mean pixel error rates for the shadow and non-shadow classes.

\subsection{Comparison with the state-of-the-art}
There is only one CNN-based video shadow detection method,TVSD \cite{chen21TVSD}. Besides, we compare our method with state-of-the-art image shadow detection methods including BDRAR \cite{Zhu_2018_ECCV}, MTMT \cite{Chen_2020_CVPR} and FSDNet \cite{hu2021revisiting}. We obtain the results of TVSD from their provided pre-trained model. We train BDRAR, MTMT, and FSDNet on the Visha dataset using their official source-codes with their default settings. 

\begin{table}[h]
\centering
\caption{\textbf{Comparison with the state-of-the-art shadow detection methods on the ViSha \cite{chen21TVSD} dataset.} 
Both Balanced Error Rate (BER) and per class error rates are shown.
Best performances are printed in bold.}
\vskip 0.1in
\label{table:quant}
\begin{tabular}{l|ccc}
\hline
Method                          &BER    &Shadow     &Non Shad.          \\
\hline
BDRAR\cite{Zhu_2018_ECCV}       &13.34  &20.80      &5.89               \\
MTMT\cite{Chen_2020_CVPR}       &14.55  &21.15      &7.94               \\
FSDNet\cite{hu2021revisiting}   &14.59  &18.88      &10.30              \\
\hline
TVSD\cite{chen21TVSD}           &16.76  &31.78      &\textbf{1.75}      \\
\hline
Ours                            &\textbf{12.02} &\textbf{11.88} &12.17  \\
\hline
\end{tabular}
\end{table}

\subsubsection{Quantitative results}
Table \ref{table:quant} reports the performance of all methods on the ViSha dataset. Our method performs best on overall BER, we obtain a 10\% error reduction and a 18\% error reduction compared with BDRAR and FSDNet, respectively.
TVSD has the best performance on non-shadow error. However, this is mainly because the method is insensitive to shadow areas, especially the fast changing shadows and the shadows with small areas. 
For the shadow areas, our method achieves the lowest error rate.
The results show that leveraging temporal information on intermediate representations can boost the shadow detection performance for videos significantly.

\begin{figure}[h!]
\includegraphics[width=\textwidth]{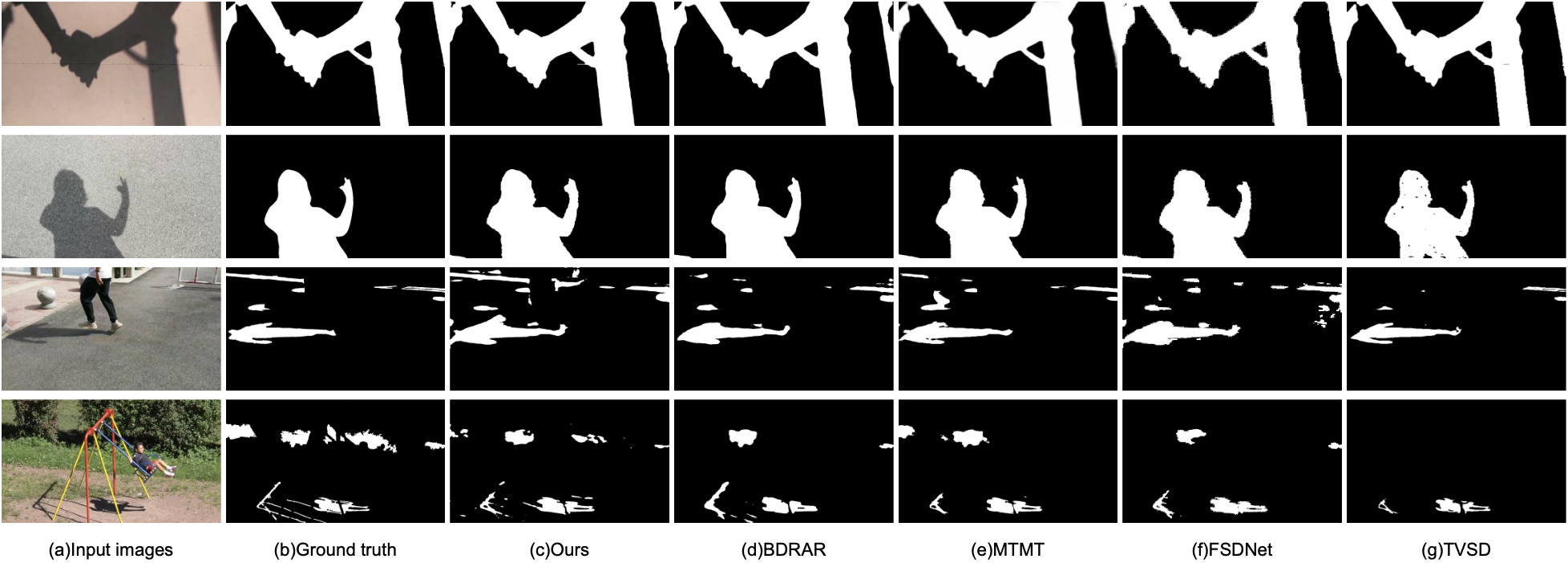}
\caption{\textbf{Comparison of shadow predictions by selected methods.} (a) shows the input images, (b) is the ground truth shadow masks and (c)-(g) are shadow masks predicted by our method and BDRAR \cite{Zhu_2018_ECCV}, MTMT \cite{Chen_2020_CVPR}, FSDNet \cite{hu2021revisiting} and TVSD \cite{chen21TVSD}.}
\label{fig:4}
\end{figure}

\begin{figure}[h!]
\centering
\includegraphics[width=\textwidth]{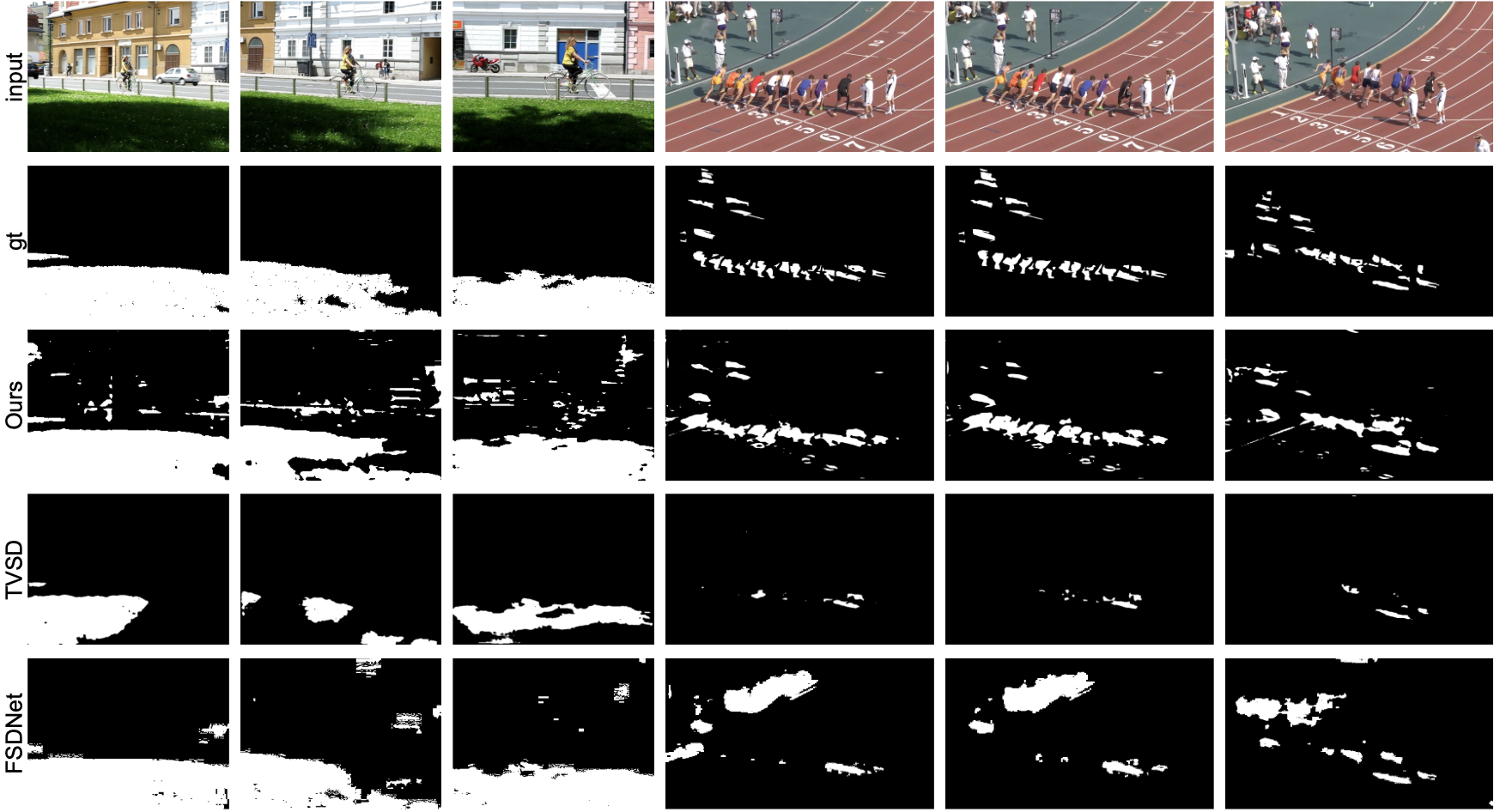}
\caption{\textbf{Video shadow detection results.} Comparison between our proposed method, FSDNet and TVSD. First row shows the input frames, second row shows the ground truth shadow masks and the bottom two are the results of comparison methods.}
\label{fig:5}
\end{figure}

\subsubsection{Qualitative results}
In Fig \ref{fig:4}, we show some shadow detection results of our method in comparison with other methods. In the first row, we can see that all methods produce fairly accurate shadow masks with clear boundaries between shadow and non-shadow areas. 
Comparing with FSDNet, our method generates shadow masks with smoother boundaries without disjoint areas.
Comparing with TVSD, our method is able to capture more shadow details, as shown in the last two rows.
Fig \ref{fig:5} shows the performance of our method on three consecutive video frames of two example videos.
As can be seen, TVSD is not robust in detecting moving shadows and not sensitive to small shadow areas. By contrast, our method predicts more accurate shadow masks in all cases. As we can see in the right three columns, only our method is able to detect the fast moving shadow regions.

\subsubsection{Failure cases}

\begin{figure}
\centering
\includegraphics[width=0.8\textwidth]{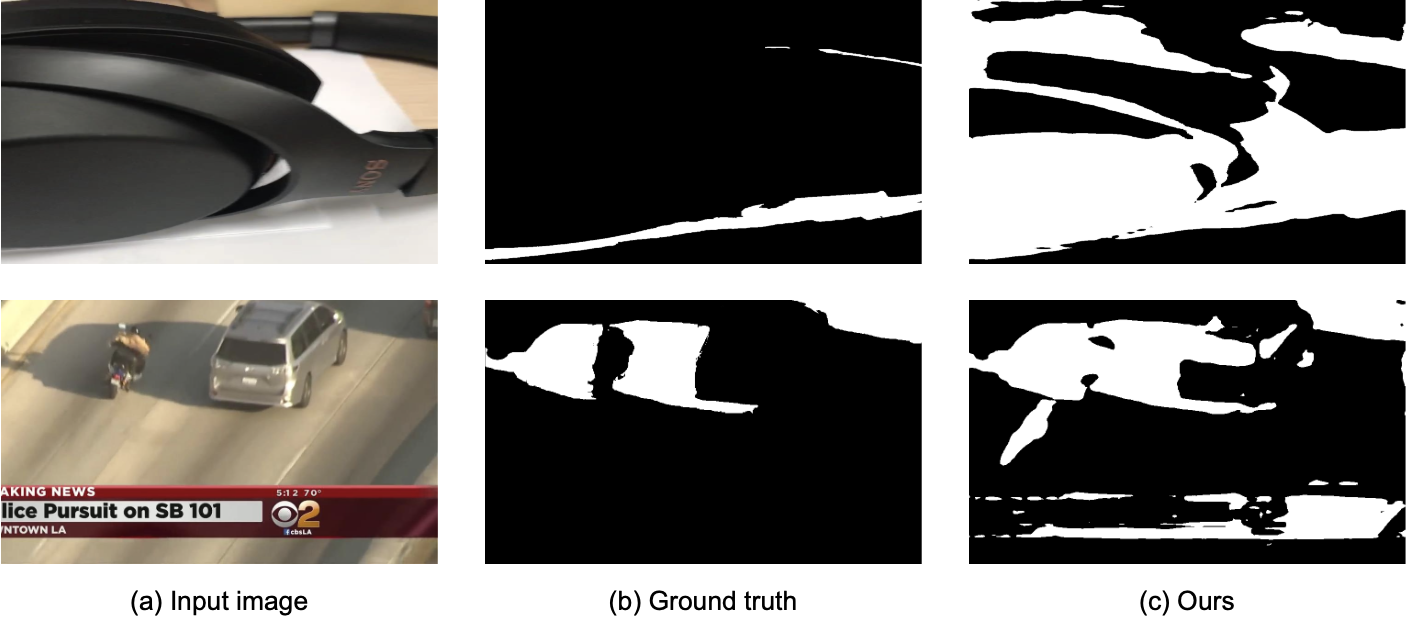}
\caption{\textbf{Failure cases.} Failure cases caused by non-shadow dark objects and limited training data diversity. (a) shows the input image, (b) shows the ground truth and (c) shows the shadow mask predicted by our method.}
\label{fig:6}
\end{figure}

Some failure cases of our method are shown in Fig \ref{fig:6}. Many of them are caused by dark objects being incorrectly classified as shadows.
Dark objects are naturally difficult for shadow detection, especially when the dark object is the majority of scene. Correctly classifying these cases requires contextual understanding of the scenes as well as sufficient training data. Note that 
the training set of the Visha dataset has only 50 videos. Improving the generalization of models trained on such small-sized dataset is challenging. 

\subsection{Application on Video Shadow Removal}
\begin{figure}[ht!]
\centering
\includegraphics[width=\textwidth]{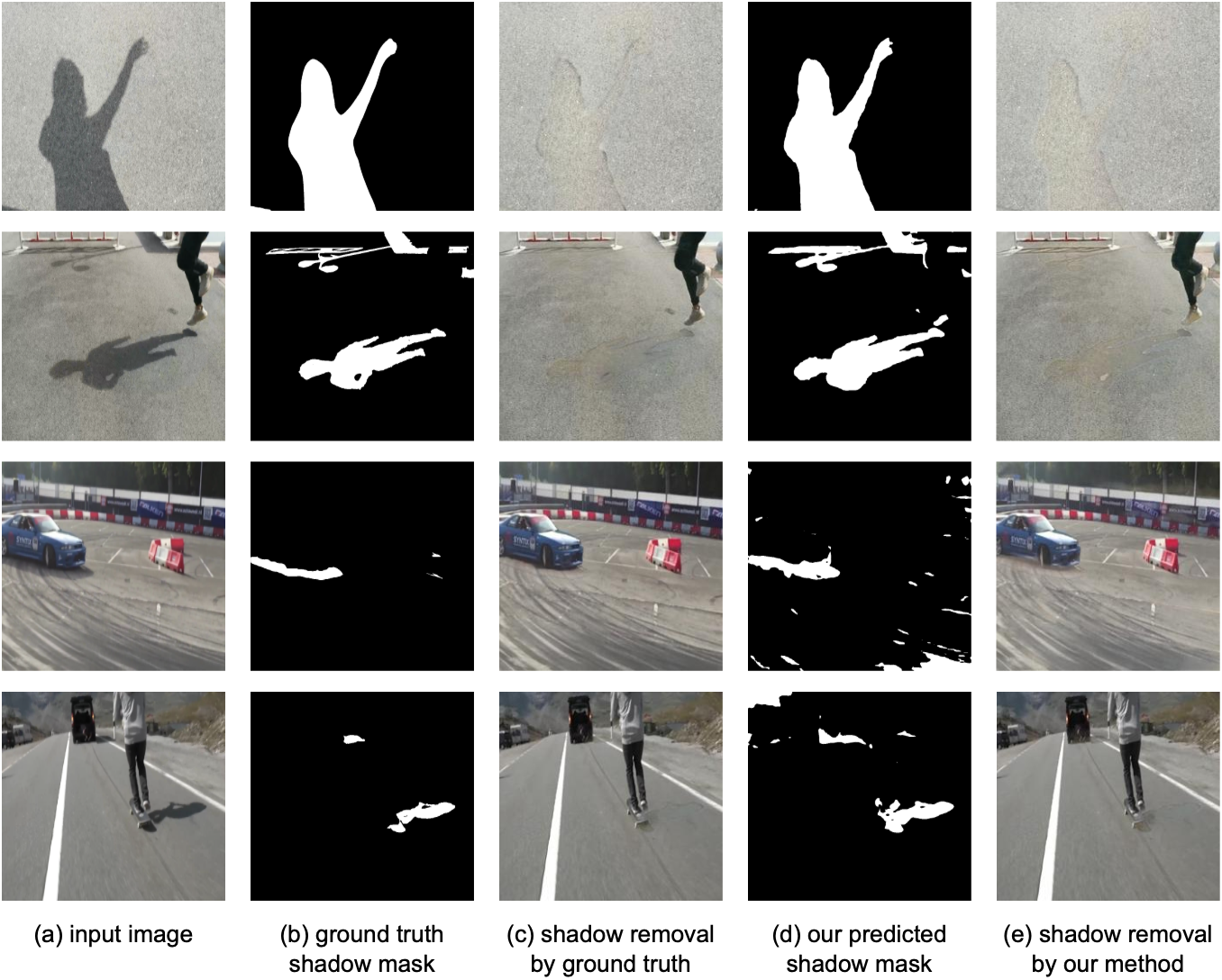}
\caption{\textbf{Application on shadow removal model.} Shadow-free image generation using shadow masks predicted by our method versus the ground truth. (a) shows the input image, (b) shows the ground truth shadow mask, (c) shows the shadow removal results using ground truth, (d) shows the shadow mask predicted by our method and (e) shows the results using our shadow mask.}
\label{fig:7}
\end{figure}

Shadow masks can be further used to remove  shadows to create shadow-free images. We test our predicted shadow masks on the model proposed in \cite{Le_2019_ICCV,le2020physicsbased} which takes the original image and the shadow mask as input and produces the shadow-free version of the image. 
Fig \ref{fig:7} shows the performance of using our predicted shadow mask and using the ground truth shadow mask, results show that our video shadow detection method gets reasonable results and can be used downstream to create shadow-free videos.

\section{Summary}
\label{sec:summary}
Large-scale video shadow detection is still in an early stage. In this paper, we show that a basic and simple idea  can be extremely effective for the task. 
We deployed an optical-flow based feature warping and combination scheme to enforce correspondence between representations.
Highly correlated intermediate representations led to an improvement of shadow prediction accuracy and consistency in the videos.
In experimental results  our model was able to handle large shadow appearance changes and capture small shadow regions, and outperformed the state-of-the-art methods on the existing video dataset.
We believe our method will serve as an important stepping stone for future work.

\bibliography{egbib,bib2}
\end{document}